\documentclass[10pt,twocolumn,letterpaper]{article}

\usepackage{iccv}
\usepackage{times}
\usepackage{epsfig}
\usepackage{graphicx}
\usepackage{amsmath}
\usepackage{amssymb}

\usepackage{bm}
\usepackage{algorithm}
\usepackage{algpseudocode}
\usepackage{graphics}
\usepackage{amsthm}
\usepackage{mathrsfs}
\usepackage{booktabs}
\usepackage{subfigure}
\usepackage{graphicx}
\usepackage{array}
\usepackage{multirow}
\usepackage{diagbox}
\usepackage[table,xcdraw]{xcolor}

\usepackage[pagebackref=true,breaklinks=true,letterpaper=true,colorlinks,bookmarks=false]{hyperref}

\newcommand{\ourmethod}{DenseTNT}

\iccvfinalcopy 

\def\iccvPaperID{7610} 

\begin{document}

\title{DenseTNT: Waymo Open Dataset Motion Prediction Challenge $1^{st}$ Place Solution}

\author{Junru Gu~~~~~~Qiao Sun~~~~~~Hang Zhao\\
IIIS, Tsinghua University\\
{\tt\small \{gujunru123,larksq,zhaohang0124\}@gmail.com}

}

\maketitle

\ificcvfinal\thispagestyle{empty}\fi

\begin{abstract}

In autonomous driving, goal-based multi-trajectory prediction methods are proved to be effective recently, where they first score goal candidates, then select a final set of goals, and finally complete trajectories based on the selected goals.
However, these methods usually involve goal predictions based on sparse predefined anchors.
In this work, we propose an anchor-free model, named DenseTNT, which performs dense goal probability estimation for trajectory prediction.
Our model achieves state-of-the-art performance, and ranks $1^{st}$ on the Waymo Open Dataset Motion Prediction Challenge.
Project page is at {\small \url{https://github.com/Tsinghua-MARS-Lab/DenseTNT}}.

\end{abstract}


\section{Introduction}

Trajectory prediction is a highly challenging task in autonomous driving due to the inherent stochasticity and multimodality of human behaviors.
To model this high degree of uncertainty, some approaches predict multiple future trajectories by sampling from the distribution represented by the latent variables, \eg VAEs~\cite{cvae} and GANs~\cite{social-gan}. Other approaches generate a fixed number of trajectories but only perform regression on the closest one during training~\cite{social-gan,lanegcn,rasterization}, namely using variety loss. 
Multipath~\cite{multipath} and CoverNet~\cite{phan2020covernet} pose the problem as a classification problem by classifying over template trajectories.

\begin{figure}[tb]
\centering
\includegraphics[width=80mm]{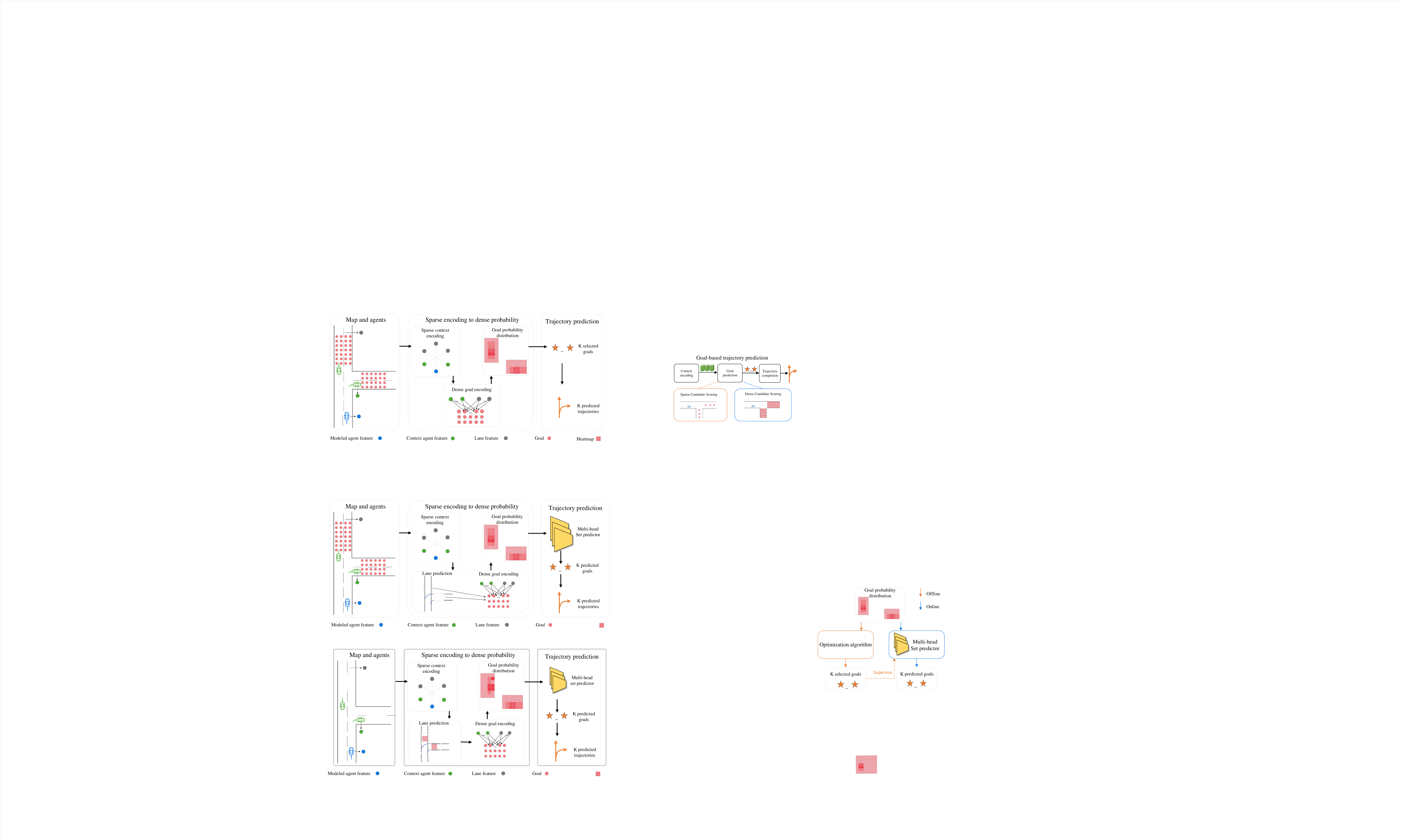}
\caption{\label{motivation}
A typical goal-based trajectory prediction pipeline is shown in the upper part of the figure.
Existing goal prediction methods~(lower left) first define sparse goal anchors heuristically, and then regress and classify these anchors to estimate the goals. In contrast, our method~(lower right) estimates the probabilities of the dense goals without relying on the quality of heuristically predefined anchors (anchor-free).
}
\end{figure}

More recently, goal-based methods~\cite{tnt,precog,Tran_2021_WACV,lanercnn} have gained popularity and achieved state-of-the-art performance. Their key observation is that the endpoint carries most of the uncertainty of a trajectory, therefore they first predicted the goal(s) of an agent, and then further completed the corresponding full trajectory for each goal.
They obtained the final goal positions by classifying and regressing predefined \textit{sparse} anchors, as shown in the lower-left part of Figure~\ref{motivation}.
For example, TNT~\cite{tnt} defined anchors as the points sampled on the lane centerlines; some others~\cite{lanercnn} took the lane segments as anchors and predicted a goal for each lane segment.

The prediction performance of these goal-based methods heavily depends on the quality of the goal anchors. Since an anchor can only generate one goal, it is impossible for the model to make multiple trajectory predictions around one anchor. Besides, different positions on the same lane segment have different local information, such as the relative distance to the nearest lane boundary. Sparse anchor-based methods cannot make use of such fine-grained information.

\begin{figure*}[tb]
\includegraphics[width=170mm]{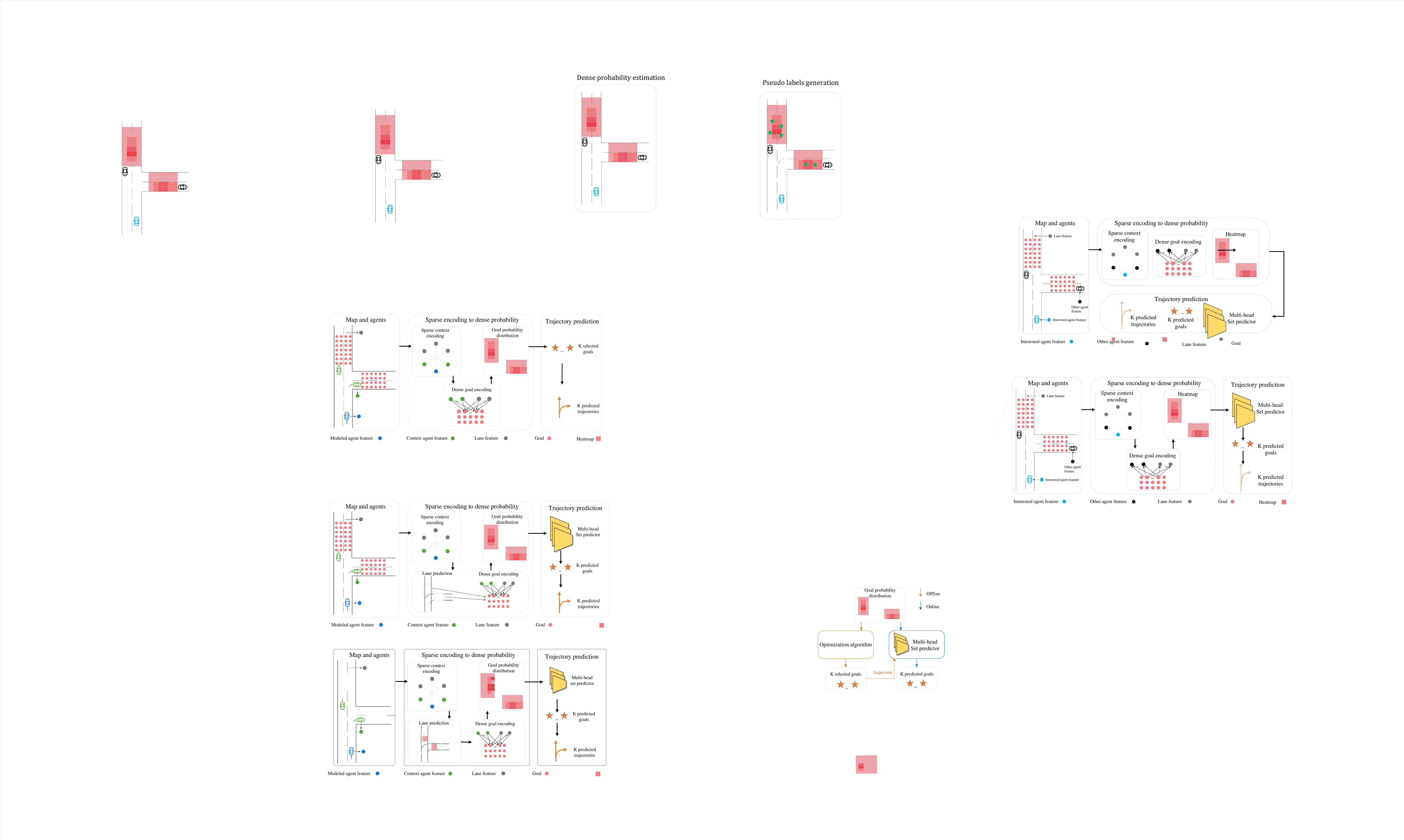}
\caption{\label{overview}
An illustration of \ourmethod. A vectorized encoding method is used to extract features of HD maps and agents; and then a dense goal encoding module is employed to output dense probability distribution.
}
\end{figure*}

In this work, we propose DenseTNT, an \textbf{anchor-free} goal-based trajectory prediction method. It largely improves the performance of goal estimation without relying on the quality of heuristically predefined goal anchors. We first extract sparse scene context features and then employ a \textit{dense} probability estimation to generate the probability distribution of the goal candidates. Finally, a trajectory completion module outputs trajectories based on a set of selected goals.

\section{Method}
\subsection{Scene context encoding}
\label{sec:sparse_encoding}

Scene context modeling is the first step in behavior prediction. It extracts the features of the lanes and the agents and captures the interactions among them.
Sparse encoding methods~\cite{vectornet,lanegcn} (also called vectorized methods) were proposed recently. Compared to dense encoding methods which rasterize the lanes and the agents into images and use CNNs to extract features, sparse encoding methods abstract all the geographic entities (\eg lanes, traffic lights) and vehicles as polylines, and better capture the structural features of high-definition maps.

\subsection{Dense goal probability estimation}
\label{sec:dense_probability}
After scene context encoding, we perform probability estimation for goals on the map. TNT~\cite{tnt} defined discretized sparse anchors on the roads and then assigned probability values upon them.
Our key observation is that sparse anchors are not a perfect approximation of real probability distributions on the roads, because (1) one anchor can only generate one goal, we cannot make multi-trajectory predictions around one anchor; (2) there are many ordinary points on the roads (those away from the lane centers or boundaries) that were not well modeled: different ordinary points on the same road have different local information, \ie the relative distance to the nearest lane boundary.

Therefore, we perform dense goal probability estimation on the map instead. Concretely, a dense goal encoding module is used to extract the features of \textit{all} the locations on the road under a certain sampling rate. Then, the probability distribution of the dense goals is predicted.

The dense goal encoding module uses an attention mechanism to extract the local information between the goals and the lanes.
We denote the feature of the $i_{th}$ goal as $F_i$, which is obtained by a 2-layer MLP, and the input of the MLP is 2D coordinates of the $i_{th}$ goal.
The local information between the goals and the lanes can be obtained by attention mechanism:
\begin{equation}
    Q = FW^Q, K = LW^K, V = LW^V 
\end{equation}
\begin{equation}
A \left( Q,K,V \right) ={\rm softmax} \left( \frac{QK^\mathrm{T}}{\sqrt{d_k}}\right)V 
\end{equation}
where $W^Q, W^K, W^V \in \mathbb{R}^{d_h\times d_k} $ are the matrices for linear projection, $d_k$ is the dimension of query / key / value vectors, and $F, L$ are feature matrices of the dense goals and all map elements (\ie, lanes or agents), respectively.

The predicted score of the $i_{th}$ goal can be written as:
\begin{equation}
\phi_i = \frac{\exp(g(F_i))}{\sum^{N}_{n=1} \exp(g(F_n))},
\end{equation}
where the trainable function $g(\cdot)$ is also implemented with a 2-layer MLP.
The loss term for training the scene context encoding and the dense probability estimation is the binary cross-entropy between the predicted goal scores and the ground truth goal scores:
\begin{equation}
\mathcal{L}_{goal} = \sum_i\mathcal{L}_{CE}(\phi_i,\psi_i),
\end{equation}
where $\psi_i$ is the ground truth score of the $i_{th}$ goal. The ground truth score of the goal closest to the final position is $1$, and the others are $0$.

\subsection{Goal selection}
After the dense probability estimation, we use non-maximum suppression (NMS) algorithm to select goals. NMS iteratively selects the goal with the highest probability and removes the goals which are close to the selected goal. The first K selected goals are the predicted goals.

\subsection{Trajectory completion}

Similar to TNT, the last step is to complete each trajectory conditioned on the selected goals.
We only have one ground truth trajectory, so we apply a teacher forcing technique~\cite{williams1989learning} by feeding the ground truth goal during training.
The loss term is the offset between the predicted trajectory $\hat{s}$ and the ground truth trajectory $s$:
\begin{equation}
\mathcal{L}_{\rm completion} = \sum\limits ^{T}_{t=1} \mathcal{L}_{\rm reg} (\hat{s}_t, s_t)
\end{equation}
where $\mathcal{L}_{\rm reg}$ is the smooth $l1$ loss between two points.

\subsection{Long-term prediction}
The previous steps can already achieve good performance in short-term (\eg 3s) motion prediction tasks. However, long-term prediction is still challenging since the probability distribution may diverge into the long future.
Inspired from sentence generation in natural language processing, we generate the probability distribution of the goals in an auto-regressive manner, at 3s, 5s and 8s respectively.

Since we aim to roll out dense probability estimation in 3 steps, we develop three branches in our model architecture. 

The three branches share the same weights for the subgraph module in scene context encoding and have independent weights for other parts, \eg the global graph module in scene context encoding and the dense probability estimation.

With $N$ goal selection at 3s, 5s and 8s auto-regressively, we obtain $N^3$ goal sets. We sort the top K goal sets according to their probability scores, and then complete them to obtain K trajectories. 

More concretely, for each goal set, we use the above dense goal encoding module to get the features of the 3 goals. Then the features are passed to the trajectory completion module that is a 2-layer MLP. The output is a full trajectory $[\hat{s}_1, \hat{s}_2, ..., \hat{s}_T]$.

\section{Experiments}
\subsection{Implementation details}
\paragraph{Agent and map encoding.}
To normalize the map, we take the last position of the target vehicle as the origin and the direction of the target vehicle as the $y$-axis. Since the map in each scenario is quite large, we only encode a submap with a center of (0, 30m) and a radius of 80m. 

Following VectorNet~\cite{vectornet}, the agents are converted into sequences of vectors. Each vector contains the start point, the end point, the timestamps of its start and end point, and the attributes of its corresponding agent. The lanes are converted into sequences of lane segments. Each lane segment contains 10 adjacent lane points and the attributes of its corresponding lane. For example, a lane of 50 lane points is converted into 5 lane segments. Since the sampling distance between two adjacent points is about 1m, a lane segment of 10 lane points is about 10m.

\paragraph{Dense goal sampling.}
Dense goal sampling aims to sample all possible goals of the target vehicle.

Only goal candidates densely located on the roads and the parking lots need to be sampled. The distance between two adjacent goals, \ie, the sampling density, is set to 1m. We don't sample the goals which are located outside of the submap defined above. 

\paragraph{Training details.}
Our model is trained on the training set with a batch size of 64. We use Adam~\cite{adam} optimizer to train 16 epochs, and the learning rate with an initial value of 0.001 is decayed by a factor of 0.3 every 5 epochs. The hidden size of the feature vectors is set to 128. No data augmentation is used, such as random perturbation or map zooming. 

There are three agent types, namely pedestrians, vehicles, and cyclists. We train a model for each of them, since different agent types have different behavioral characteristics.

\begin{table}[htb]

\centering
\caption{Comparison of sparse and dense goal probability estimation on the Argoverse Forecasting validation set.}
\begin{tabular}{c|c|c|c}
\toprule
\multicolumn{2}{c|}{Goal Probability Estimation} & \multirow{2}{*}{minFDE} & \multirow{2}{*}{Miss Rate} \\ \cline{1-2}
Sparse               & Dense                &                         &                            \\ \hline
\checkmark            &                      & 1.35                    & 9.5\%                      \\
                     & \checkmark            & 1.28                    & 8.2\%                      \\ \bottomrule
\end{tabular}

\label{ablation}

\end{table}

\begin{table*}[tb]
\centering
\caption{Top 10 entries of the Waymo Open Dataset Motion Prediction Challenge. mAP is the official ranking metric.}
\begin{tabular}{lccccc}
\toprule
Method              & minADE & minFDE & Miss Rate & Overlap Rate &\cellcolor[HTML]{EFEFEF} mAP    \\ \hline
DenseTNT $1^{st}$ (Ours)    & 1.0387 & 1.5514 &  0.1573    & 0.1779      &\cellcolor[HTML]{EFEFEF} \bf 0.3281 \\
TVN  $2^{nd}$       & 0.7558 & 1.5859 & 0.2032    &\bf 0.1467    &\cellcolor[HTML]{EFEFEF} 0.3168 \\
Star Platinum $3^{rd}$& 0.8102 & 1.7605 & 0.2341    & 0.1774     &\cellcolor[HTML]{EFEFEF} 0.2806 \\
SceneTransformer    &\bf 0.6117 &\bf 1.2116 &\bf 0.1564    & 0.1473&\cellcolor[HTML]{EFEFEF} 0.2788 \\
ReCoAt              & 0.7703 & 1.6668 & 0.2437    & 0.1642       &\cellcolor[HTML]{EFEFEF} 0.2711 \\
AIR                 & 0.8682 & 1.6691 & 0.2333    & 0.1583       &\cellcolor[HTML]{EFEFEF} 0.2596 \\
SimpleCNNOnRaster   & 0.7400 & 1.4936 & 0.2091    & 0.1560       &\cellcolor[HTML]{EFEFEF} 0.2136 \\
CNN-MultiRegressor  & 0.8257 & 1.7101 & 0.2735    & 0.1640       &\cellcolor[HTML]{EFEFEF} 0.1944 \\
GOAT                & 0.7948 & 1.6838 & 0.2431    & 0.1726       &\cellcolor[HTML]{EFEFEF} 0.1930 \\
Waymo LSTM baseline & 1.0065 & 2.3553 & 0.3750    & 0.1898       &\cellcolor[HTML]{EFEFEF} 0.1756 \\ \bottomrule
\end{tabular}
\label{leaderboard}
\end{table*}

\begin{table*}[]
\centering
\caption{Performance breakdown of DenseTNT on different categories.}
\begin{tabular}{lccccc}
\toprule
Object Type & minADE & minFDE & Miss Rate & Overlap Rate & mAP    \\ \midrule
Vehicle     & 1.3462 & 1.9120 & 0.1518    & 0.1296       & 0.3698 \\
Pedestrian  & 0.5013 & 0.9130 & 0.1014    & 0.2725       & 0.3342 \\
Cyclist     & 1.2687 & 1.8292 & 0.2186    & 0.1314       & 0.2802 \\ \bottomrule
\end{tabular}
\label{agent_type}
\end{table*}

\begin{figure*}[!htb]
\centering
\includegraphics[width=170mm]{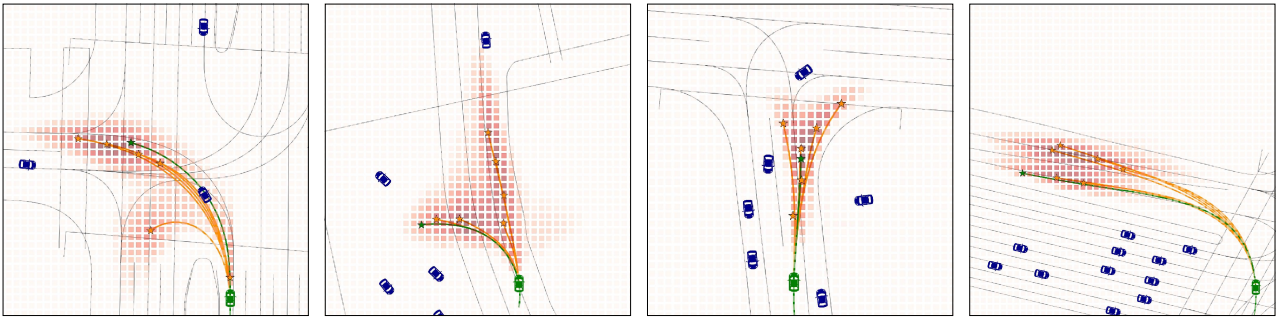}
\caption{Qualitative results of DenseTNT on the Waymo Open Motion Prediction validation set. The probabilities of the dense goals are shown in red, selected goals and corresponding trajectories are shown in orange, ground truth trajectories are shown in green.
}
\label{visual}
\end{figure*}

\subsection{Results}
\paragraph{Sparse and dense goals.}
We compare and evaluate sparse and dense goal probability estimation on the Argoverse Forecasting dataset. As shown in Table~\ref{ablation}, the dense model gives better performance than the sparse model, which is vanilla TNT.

\paragraph{Waymo Open Dataset Motion Prediction Challenge.}
We evaluate the effectiveness of DenseTNT on the Waymo Open Dataset Motion Prediction Challenge. As shown in Table~\ref{leaderboard}, our method ranks $1^{st}$ on the leaderboard. The official metric is mAP, which provides a full picture of the model performance~\cite{waymo}.
The breakdown performance on each category is shown in Table~\ref{agent_type}.

\subsection{Qualitative results}

In DenseTNT, the goal candidates are densely distributed on the map. We visualize the probabilities of the dense goals and the predicted trajectories based on the selected goals. As shown in Figure~\ref{visual}, DenseTNT gives diverse predictions such as going straight, left/right turns and U-turns.

\section{Conclusion}
In this report, we propose an anchor-free trajectory prediction model, named DenseTNT. It outperforms previous goal-based methods by removing heuristically predefined goal anchors. DenseTNT achieves state-of-the-art performance and ranks $1^{st}$ on the Waymo Open Dataset Motion Prediction Challenge.

{\small
\bibliographystyle{ieee_fullname}
\bibliography{egbib}
}

\end{document}